\documentclass{article}

\usepackage{graphicx} 
\usepackage{amsmath}
\usepackage{amssymb} 
\usepackage{dsfont}
\usepackage[ruled,Algoritm]{algorithm}
\usepackage{algorithmic}
\usepackage{booktabs} 
\usepackage{url}

\title{A Probabilistic Model
for Node Classification in Directed Graphs}
\author{Diego Huerta and Gerardo Arizmendi}
\date{\today}

\begin{document}

\maketitle

\begin{abstract}
In this work, we present a probabilistic model 
for directed graphs where nodes have
attributes and labels. 
This model serves as a generative classifier
capable of predicting the labels of unseen nodes
using either maximum likelihood or 
maximum a posteriori estimations.  
The predictions made by this model 
are highly interpretable, 
contrasting with some common methods
for node classification,
such as graph neural networks.
We applied the model to two datasets, 
demonstrating predictive performance 
that is competitive with, 
and even superior to,
state-of-the-art methods.
One of the datasets considered is adapted 
from the Math Genealogy Project, 
which has not previously been utilized for this purpose. 
Consequently, we evaluated several 
classification algorithms
on this dataset to compare the performance 
of our model and provide benchmarks for this new resource.
\end{abstract}

\section{Introduction}


A graph is a pair $G = (V, E)$ 
representing a collection of entities $V$,
called nodes, and a binary relation between 
them $E \subset V \times V$.
This general structure enables graphs to effectively 
represent a diverse array of objects across
various domains,
including physics, biology, social networks,
and chemistry, among others.
Consequently, in recent years,
there has been a substantial increase 
in the availability and significance 
of graph-structured data \cite{network_book, dl_graphs}.
Within this context, 
the ability to learn from this graph representation
becomes essential. 
Specifically, given a property or label 
available for a subset of the nodes,
a common task is to extend this property
to unlabeled nodes. 
This task, known as node classification,
has gained importance in a variety of fields \cite{node_class_online, node_class_sn}.

For node classification,
machine learning techniques, 
especially Graph Neural Networks (GNNs),
are widely used due to 
their ability to capture relational 
information in graph-structured data \cite{graph_representation_learning_book, gnn_book}.
GNNs have achieved state-of-the-art 
performance across various tasks,
making them a preferred choice for node classification \cite{gnn_survey, gcn_review}.
However, a major challenge for neural network based models 
is their lack of interpretability.
Despite high predictive accuracy,
their complex architectures often 
lead to opaque decision-making processes,
rendering them ``black boxes" 
\cite{ml_interpretable_survey, explainable_ml}.
Understanding a model's decision-making 
is essential for trust in its predictions
\cite{explainable}, 
and in some cases, 
simpler interpretable models are favored 
\cite{interpretability_vs_explain}.
This is particularly crucial in fields such as 
healthcare, finance, and criminal justice, 
where transparency and trustworthiness are vital 
\cite{interpretable_healthcare, interpretability_finance, interpretable_criminal}.

In response to this,
we propose a non-neural network model
for node classification 
that yields interpretable predictions. 
Specifically, we introduce a 
probabilistic model to describe
the underlying behavior of the data
presented in a directed graph.
This general model allows us to define
the degree and label distribution, 
as well as the behavior of labels of connected nodes.
Namely, the model defines 
the probabilities associated with several events, 
taking into account the information within the graph. 
This model serves as a 
generative classifier for node classification, 
where we consider two approaches for making predictions:
maximum likelihood estimations 
and maximum a posteriori estimations.
To estimate the probability
regarding the label of a single node ,
the proposed model considers information 
from the first-order neighborhood 
and assumes conditional
independence among each component.
Making inferences using the proposed model
is not computationally expensive,
and the iterative prediction process
can even be parallelized 
to enhance execution time.
Moreover, by estimating the parameters
of the probabilistic model,
we can make predictions for unseen nodes in the graph. 
This means that not all nodes in the graph need 
to be present during training, 
allowing us to obtain an inductive method 
for node classification \cite{graphsage}.

We applied our model 
for node classification on two datasets,
comparing its performance with benchmarks and
other methods for node classification,
including Graph Neural Networks.
The first dataset for testing our method 
is a newly created dataset derived 
from the publicly available information
of the Math Genealogy Project\footnote{\url{https://mathgenealogy.org/index.php}},
specifically adapted for the node classification task.
This data has never been used before 
for classification tasks, 
necessitating adaptations for this purpose. 
We also evaluated 
several common classification methods 
on this dataset to serve as benchmarks
and to compare their predictions with
those generated by our model.
Additionally, we applied our model 
to the ogbn-arxiv dataset \cite{ogbn_arxiv_paper}, 
a widely recognized dataset used 
for benchmarking node classification algorithms.

The contributions of this work are:
\begin{itemize}
    \item Introduction of a probabilistic model for describing directed attributed graphs.
    \item Utilization of this model for node classification, resulting in interpretable predictions. All the code is available in a GitHub repository.\footnote{
    \url{https://github.com/DiegoHuerta1/A-Probabilistic-Model-for-Node-Classification-in-Directed-Graphs}}
    \item Introduction of a new dataset for node classification, consisting of an adaptation of the Math Genealogy Project data.
\end{itemize}


\section{Preliminaries}

This section introduces the foundational concepts 
necessary for describing
our model and 
the experimental evaluation of node classification.
We begin by establishing notation in probability theory
and introducing parametric distribution families 
relevant to our experiments.
Following this, we provide an overview of
several standard machine learning
classification algorithms,
which are subsequently evaluated in the experiments.

\subsection{Probability theory}

Let $Z$ be an $m$-dimensional discrete random vector
($m=1$ for the case of a random variable),
we denote by
$f_Z:\mathbb{R}^m \rightarrow [0, 1]$ 
the probability mass function of $Z$,
given by

\begin{equation*}
    f_Z (z) = \mathbb{P} (Z = z),
\end{equation*}

for each $z \in \mathbb{R}^m$.
For two discrete random vectors $Z$ and $Z'$,
of dimensions $m$ and $m'$, respectively,
we denote by
$f_{Z, Z'}:\mathbb{R}^{m + m'} \rightarrow [0, 1]$ 
their joint probability mass function.
It is given by 

\begin{equation*}
    f_{Z, Z'} (z, z') = \mathbb{P} (Z = z, Z' = z'),
\end{equation*}

for each $z \in \mathbb{R}^m$, $z' \in \mathbb{R}^{m'}$.
For two random variables (or vectors) 
$Z$ and $Z'$,
the conditional probability function of
$Z$ given $Z' = z'$, for $z'$ such that $f_{Z'}(z') > 0$,
is expressed as

\begin{equation*}
f_{Z|Z'} (z, z') = \frac{f_{Z,Z'} (z, z')} {f_{Z'} (z')}.
\end{equation*}

\subsubsection{Multinomial distribution}

Let $d, k \in \mathbb{N}$, and let
$\theta \in \mathbb{R}^k$ be a vector
$\theta = (\theta_1, \theta_2, \dots, \theta_k)$
such that
$\theta_i \geq 0$ for each 
$i \in \{1, \dots, k\}$,
and 
$\sum_{i=1}^k \theta_i = 1$.
A $k$-dimensional discrete random vector $Z$
is said to follow a multinomial distribution
with parameters $d$ and $\theta$,
denoted by $Z \sim \text{Multinomial}(d, \theta)$,
if

\begin{equation*}
    f_Z (z) = 
    \begin{cases}
    \frac {d!}{\prod_{i=1}^k z_i!} 
    \prod_{i=1}^k (\theta_i)^{z_i}
    & \text{if }  \sum_{i=1}^k z_i = d,\\
    0 & \text{otherwise.}
\end{cases}
\end{equation*}

for all $z = (z_1, z_2, \dots, z_k) \in (\mathbb{N}_0)^k$,
where $\mathbb{N}_0 = \{0, 1, 2, \dots\}$
denotes the set
of non-negative integers. 
Denote by 
$g(z; d, \theta)$ the 
probability mass function
of a discrete random vector
that follows a multinomial distribution
with parameters $d$ and $\theta$,
evaluated at $z$.

\subsubsection{Discrete truncated power law}

A discrete random variable $Z$
with support on the positive integers
is said to follow a power law distribution
with parameter $\kappa > 0$
if its probability mass
function $f_Z$ satisfies

\begin{equation*} 
    f_Z(z) \propto z^{-\kappa}.
\end{equation*}

The truncated power law distribution
(also known as power law with exponential cutoff),
is a power law multiplied 
by an exponential function.
That is,
the discrete random variable
$Z$
with support on the positive integers
is said to follow a truncated
power law distribution,
with parameters $\kappa > 0$,
and $\lambda > 0$
if its probability mass
function satisfies

\begin{equation*}
    f_Z(z) \propto z^{-\kappa}
    e ^{-z\lambda}.
\end{equation*}

\subsubsection{Log-normal distribution}

A continuous random variable $Z$ 
is said to follow a log-normal distribution 
with parameters $\mu$ and $\sigma^2$,
denoted as $Z \sim \text{Lognormal}(\mu, \sigma^2)$,
if the natural logarithm of $Z$ 
follows a normal distribution with
mean $\mu$ and variance $\sigma^2$, 
i.e., $\ln(Z) \sim \mathcal{N}(\mu, \sigma^2)$.
The probability density function 
$f_Z(z)$ of a log-normal random variable is given by

\begin{equation}\label{lognormal_pdf}
f_Z(z) = (\sigma z \sqrt{2 \pi})^{-1} 
\text{exp}\left( - \frac{(\text{ln}(z) - \mu)^2}{2 \sigma^2} \right).    
\end{equation}

The log-normal distribution is widely used
for modeling citation count data,
which is inherently discrete in nature
\cite{discrete_lognormal_citation_data}.
However, 
applying the log-normal
distribution to such data 
requires a discretized version 
of the distribution.
Several approaches exist for discretizing
the continuous log-normal distribution.
For example, one common method involves
calculating the probability of a discrete 
value by integrating the continuous 
probability density function 
over the unit interval surrounding that value.
In this work, 
we adopt the discretization approach
described in \cite{discrete_lognormal}.
Specifically, we treat the p.d.f. given 
in equation (\ref{lognormal_pdf})
as the probability density function 
of a discrete log-normal random variable,
with the appropriate normalization.

\subsection{Machine Learning Models for Classification}

This section describes three 
common machine learning methods 
used for classification.
Specifically, we outline:
the Naive Bayes algorithm,
a probabilistic classifier;
BERT, a neural network-based model 
that can be used for text classification;
and Graph Convolutional Networks (GCNs),
a neural network-based approach for node classification in graphs. 
These methods are selected for their
diverse theoretical foundations and 
proven performance across various applications.
We apply these three methods to the
Math Genealogy Project dataset
to evaluate and compare the effectiveness
of our probabilistic model in classification.

\subsubsection{Naive Bayes} \label{Naive bayes}

The Naive Bayes algorithm
is a well-known 
and efficient probabilistic classifier 
with a wide range of applications \cite{NB_aplications}.
It is frequently employed in text classification
due to its computational efficiency
and relatively strong predictive performance,
often competitive with more sophisticated methods 
\cite{NB_text_feature, NB_bayesian}.
Here, we describe the Naive Bayes classifier 
in the context of text classification,
following the detailed explanation in Chapter 4 of \cite{naive_bayes_text_book}.

Multinomial Naive Bayes is a 
supervised generative classification algorithm
that assigns documents to predefined classes 
based on their content. 
The classifier is trained on a dataset 
$\{(x_i, y_i)\}_{i=1}^N$, 
where each $x_i$ represents a document
and $y_i \in \mathcal{Y}$ is
its corresponding class label.
Each document $x_i$ is represented by
the terms (i.e. words) $t_j$ it contains,
where the algorithm assumes a bag-of-words 
representation
\cite{bag_of_words}
(i.e., word identity matters,
but not their position within the document). 
Given a new document,
with terms $t_1, \dots, t_m$,
the Naive Bayes classifier predicts 
the class by maximizing the posterior probability
of the class given the terms 
in the document, as follows:

\begin{equation*}
    y_{\text{NB}} = \arg\max_{y \in \mathcal{Y}} 
    \left( \log \mathbb{P}(y) + 
    \sum_{j \in \text{positions}} \log \mathbb{P}(t_j \,| \,y) \right).
\end{equation*}

Here, $\mathbb{P}(y)$ represents 
the prior probability of class $y$,
and $\mathbb{P}(t_j \,|\, y)$ represents
the likelihood of the term $t_j$ 
appearing in a document of class $y$.
To estimate the prior probabilities $\mathbb{P}(y)$ 
using maximum likelihood, 
let $N_y$ be the number of documents 
in the training data with class $y$,
and $N$
be the total number of documents.
The estimates of the prior  probabilities
are given by:

\begin{equation*}
    \hat{\mathbb{P}}(y) = \frac{N_y}{N}.
\end{equation*}

For the likelihood of a term $t_j$ given a class $y$,
Laplace smoothing can be employed on word frequencies
\cite{laplace_smoothing}.
This smoothing addresses the problem of
zero probabilities for words that do not appear
in the training data for a specific class.
The likelihood $\mathbb{P}(t_j \,| \,y)$ 
is estimated by concatenating 
all documents of class $y$,
and using the frequency of the term $t_j$ 
in this concatenated document. 
The Laplace-smoothed estimate is given by:

\begin{equation} \label{eq_naive_bayes_proba_palabra}
    \hat{\mathbb{P}}(t_j \,| \, y) = 
    \frac{ \text{count}(t_j, y) + 1 }
    {\sum_{t \in V} \left( \text{count}(t, y) + 1 \right)}.
\end{equation}

where $V$ is the vocabulary consisting 
of all possible terms in the dataset.
The smoothing term ``1" in both
the numerator and denominator 
of Equation \ref{eq_naive_bayes_proba_palabra} 
can be replaced by a hyperparameter $\alpha$ 
to generalize 
the estimation method to
to additive smoothing
\cite{additive_smoothing}.

\subsubsection{BERT}

Bidirectional Encoder 
Representations from Transformers (BERT) 
is a language representation model 
designed to pretrain deep bidirectional
representations from unlabeled text \cite{bert_original}.
BERT has been successfully applied to a
variety of Natural Language Processing (NLP) tasks
\cite{bert_nlp},
such as question answering \cite{bert_original},
sentiment analysis \cite{bert_sentiment},
depression classification \cite{bert_depresion},
and phishing email detection \cite{bert_email}.

BERT is built on the transformer model,
a neural network model based 
on attention mechanisms \cite{transformer_original}.
Its architecture is a multi-layer bidirectional
Transformer encoder,
comprising $L = 12$ layers,
a hidden size of $H = 768$,
and $A = 12$ self-attention heads,
contributing to a total of 110 million parameters \cite{bert_original}.
The input to BERT is a sequence of text,
to which a special token,
called the classification token
and denoted [CLS], 
is added at the beginning.
The model outputs vector representations
$T_i \in \mathbb{R}^H$
for each term in the sequence,
along with a representation 
$C \in \mathbb{R}^H$
for the [CLS] token.
This [CLS] vector serves as 
an aggregate representation for classification tasks.
BERT is pretrained using 
the Masked Language Model (MLM) 
and Next Sentence Prediction (NSP) objectives,
which allow it to learn general language representations.

For classification tasks,
BERT can be utilized by 
initializing the model with
pretrained parameters and appending 
a classification layer. 
This output layer takes the vector
associated with the [CLS] token, $C$, 
and computes the predicted class. 
The classification layer introduces
a new parameter matrix 
$W \in \mathbb{R}^{K \times H}$,
where $K$ is the number of distinct classes
in the classification problem.
All model parameters, 
including those from BERT,
are fine-tuned during the training phase. 
Given the vector $C$, 
the output layer produces 
the probabilities for the $K$ classes,
represented by $Z \in \mathbb{R}^K$,
by computing:

\begin{equation*}
Z =  \text{softmax}(C W^T).
\end{equation*}

\subsubsection{Graph Convolutional Networks}

Graph Convolutional Networks (GCNs)
are neural network-based models
designed for classifying nodes within a graph,
where labels are only available
for a subset of the nodes 
\cite{gcn_original}.
GCNs have demonstrated strong capabilities
in learning effective graph representations,
resulting in superior performance
across a variety of tasks and applications
\cite{gcn_review}.
They have been 
successfully applied in diverse fields,
including computer vision 
\cite{gcn_cv_cnn, gcn_vision},
natural language processing 
\cite{gcn_nlp, gcn_abusive_language},
and anomaly detection \cite{gcn_anomaly}. 
Their versatility in handling different data types
and expressive power makes them 
a compelling choice for node classification tasks.

Let $G = (V, E)$ be a graph with $n = |V|$ nodes,
an adjacency matrix 
$A \in \mathbb{R}^{n \times n}$,
and a feature matrix 
$X \in \mathbb{R}^{n \times d}$,
where each row of $X$ represents 
the $d$-dimensional feature vector
associated with a node.
The ordering of nodes in $X$ 
is consistent with that in $A$.
Additionally, each node $i$ 
in the training set $V_{\text{train}} \subseteq V$ 
has a known label $y_i$ 
out of $C$ possible classes.
The objective is to predict labels
for nodes outside of the training set.
A general Graph Neural Network (GNN) model 
for node classification 
\cite{gnn_book, graph_representation_learning_book}
is composed of $L$ graph filtering operations (layers)
followed by a final classification layer.
Each layer updates node features based on 
the features from the previous layer
and information from each node's neighborhood.
Formally, 
the $l$-th graph filtering layer $h_l$ 
takes the form

\begin{equation*}
    X^{(l)} = h_l(A, X^{(l-1)}),
\end{equation*}

where $X^{(l)} \in \mathbb{R}^{n \times d^{(l)}}$
represents the node features at the $l$-th layer,
with each node having a 
$d^{(l)}$-dimensional feature vector.
The initial features are 
$X^{(0)} = X$ and $d^{(0)} = d$.
In the Graph Convolutional Network (GCN) model
\cite{gcn_original}, 
each layer $h_l$ is defined as

\begin{equation*}
    X^{(l)} = h_l(A, X^{(l-1)})
    = \sigma
    \left(
    \tilde{D}^{-\frac{1}{2}} \tilde{A} \tilde{D}^{-\frac{1}{2}}
     X^{(l-1)} W^{(l)}
    \right).
\end{equation*}

Here, $\tilde{A} = A + I_n$ 
is the adjacency matrix with added self-loops,
$I_n \in \mathbb{R}^{n \times n}$ 
is the identity matrix,
and $\tilde{D} \in \mathbb{R}^{n \times n}$
is the degree matrix of $\tilde{A}$,
that is $\tilde{D}_{ii} = \sum_{j=1}^n \tilde{A}_{ij}$.
The matrix $W^{(l)}$ represents 
the learnable weights for layer $l$,
and $\sigma$ is an element-wise 
non-linear
activation function. 
The output of the $L$-th graph filtering layer
provides the final node features $X^{(L)}$,
which are used for node classification as follows

\begin{equation*}
    Z = \text{softmax}(X^{(L)}W).
\end{equation*}

In this equation, 
the softmax function 
is applied row-wise to $X^{(L)} W$,
where 
$W \in \mathbb{R}^{d^{(L)} \times C}$ 
is a learnable parameter matrix.
This transformation maps the final node features $X^{(L)}$
to the output probabilities
$Z \in \mathbb{R}^{n \times C}$,
with each row $Z_i$ 
representing the probabilities of different classes
for the $i$-th node.
The entire GCN model can be expressed as

\begin{equation*}
    Z = f_{GCN}(A, X, \Theta),
\end{equation*}

where $\Theta$ includes 
the final matrix $W$ and
the weight matrices $W^{(l)}$ 
for each layer.
These parameters are learned by minimizing 
the cross-entropy loss function $\ell(.,.)$
over all training nodes. Namely, by minimizing

\begin{equation*}
    \mathcal{L} = \sum_{i \in V_{\text{train}}}
    \ell(Z_i,y_i).
\end{equation*}

\section{Model} \label{modelo}

\begin{figure}[ht]
    \centering
    \includegraphics[width=0.5\linewidth]{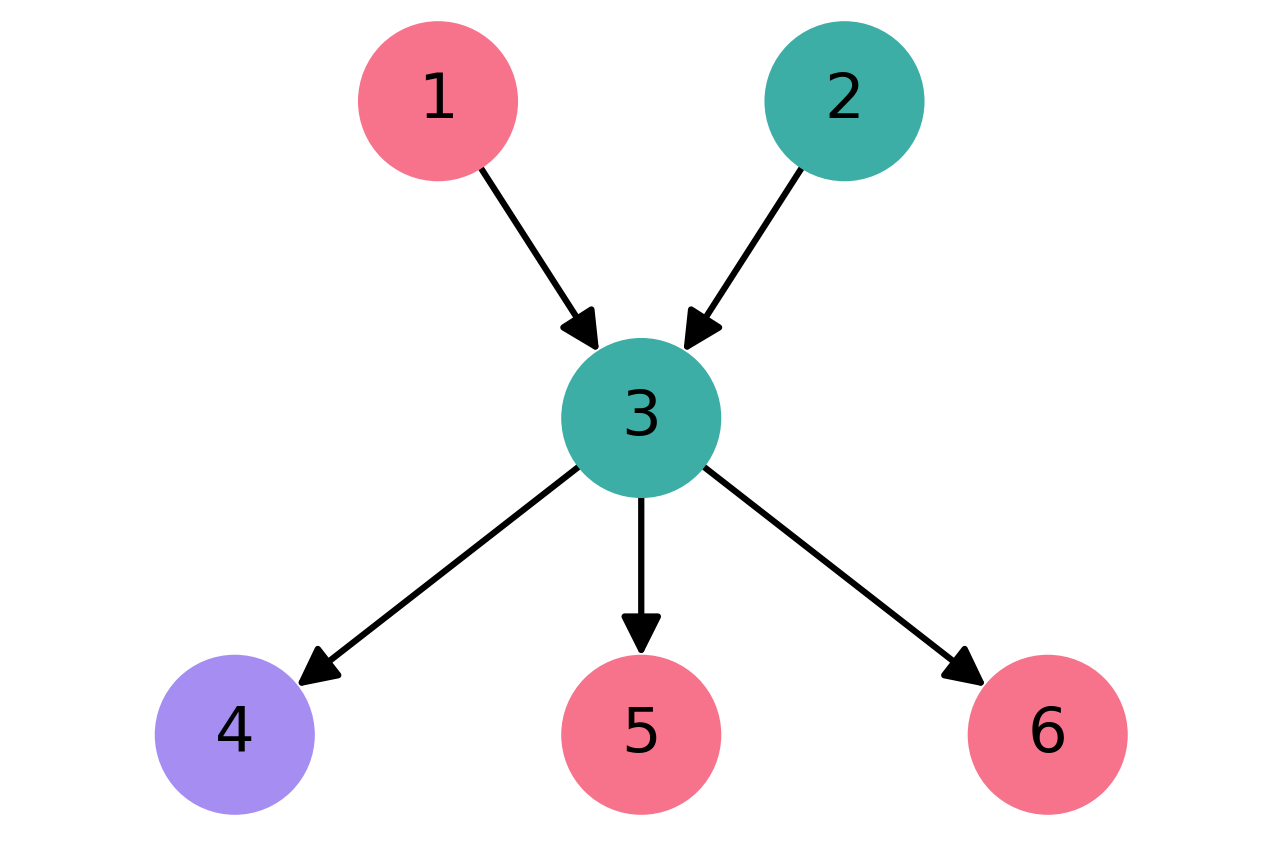}
    \caption{Example of notation. Let the
    label of a node denote its color, and consider $\mathcal{Y} = \{ 1, 2, 3\}$
    where label 1 indicates red, 
    label 2 indicates green, and
    label 3 indicates purple. Therefore,
    $y_3 = 2$, 
    $N^\text{in}(3) = \{1, 2\}$,
    $N^\text{out}(3) = \{4, 5, 6\}$
    $d_3^{\text{in}} = 2$,
    $d_3^{\text{out}} = 3$,
    $p_3 = (1, 1, 0)$ and
    $s_3 = (2, 0, 1)$.}
    \label{fig:ejemplo_notacionl}
\end{figure}

Consider a directed simple graph
$G = (V, E)$,
with $n = |V|$ nodes, 
where $V = \{1, 2, \dots,  n\}$
and $m = |E|$ edges,
where $E \subset V \times V$.
For each node $v \in V$,
$x_v \in \mathcal{X}$
denotes the attributes of the node,
and
$y_v \in \mathcal{Y}$ represents the label assigned to node $v$.
We assume that the attribute set
$\mathcal{X}$ is any countable set,
and the label set
$\mathcal{Y}$ contains
$K$ distinct labels,
$\mathcal{Y} = \{1, 2, \dots, K\}$.
Let $v \in V$ be a node in the graph,
denote by $N^\text{in}(v)$ and $N^\text{out}(v)$
the sets of predecessors and successors of $v$.
That is,
$N^\text{in}(v) = \{ u \in V : (u, v) \in E \}$
and
$N^\text{out}(v) = \{ u \in V : (v, u) \in E \}$,
the neighborhood of a node $v$ is given by
$N(v) = N^\text{in}(v) \cup N^\text{out}(v)$.
The in-degree $d_v^{\text{in}}$
and out-degree $d_v^{\text{out}}$
of a node $v$ are defined as
$d_v^{\text{in}} = |N^\text{in}(v)|$ and
$d_v^{\text{out}} = |N^\text{out}(v)|$.
Moreover, 
let
$p_v \in (\mathbb{N}_0)^K$
and
$s_v \in (\mathbb{N}_0)^K$
represent the vectors indicating the frequencies
of labels in the
predecessors
and successors
of $v$, respectively.
That is,

\begin{align*}
(p_v)_k & = |\{ u \in N^\text{in}(v) : y_u = k\}|, \\
(s_v)_k & = |\{ u \in N^\text{out}(v) : y_u = k\}|.
\end{align*}

for each $k \in \mathcal{Y}$.
Note that
$\sum_{k=1}^K (p_v)_k = d_v^{\text{in}}$
and
$\sum_{k=1}^K (s_v)_k = d_v^{\text{out}}$.
In Figure \ref{fig:ejemplo_notacionl}
we provide a simple example that lustrates this notation.

For every node $v \in V$,
denote by 
$X_v$ and $Y_v$ the random variables
indicating the attributes and the label of $v$,
and denote by 
$P_v$ and 
$S_v$ the 
random vectors indicating the 
frequencies of labels in the predecessors and successors of $v$.
We assume that these random variables
follow the same distribution
for every node $v \in V$.
Therefore, any differences
in the probabilities associated
with distinct nodes 
arise from conditioning on different events.
We assume that, 
conditioned on every other random variable of the graph,
$Y_v$ is only dependent
on $P_v$, $S_v$ and $X_v$.
That is, the label of a node only depends on 
the labels of the direct
predecessors and successors of the node,
and on its own attributes.
Additionally, we assume that 
these random variables 
are conditionally independent given $Y_v$,
namely

\begin{align} \label{decomposition}
    f_{P_v, S_v, X_v | Y_v} (p, s, x, i) = 
    f_{P_v | Y_v} (p, i)
    f_{S_v | Y_v} (s, i)
    f_{X_v | Y_v} (x, i).
\end{align}

Next, we describe the probabilistic behavior governing the labels.
Denote by 
$\pi = (\pi_1, \pi_2, \dots, \pi_K) \in \mathbb{R}^K$
the vector containing the
information of the probability mass function
of the marginal distribution of  $Y_v$.
Formally, 

\begin{equation}
    \pi_i = f_{Y_v}(i) = \mathbb{P} (Y_v = i),
    \quad \forall i \in \mathcal{Y}.
\end{equation}

The labels of nodes in the neighborhood of $v$
directly impact these probabilities.
For instance,
let $\Theta \in \mathbb{R}^{K \times K}$
be a matrix with the conditional probabilities
of a node label conditioned on a predecessor label.
More specifically,
let $\Theta_{i, j}$ 
represent the probability that a node has label $j$,
given that one of its predecessors has label $i$.
Analogously, 
let $\Xi \in \mathbb{R}^{K \times K}$
be a matrix with conditional probabilities
of a node label conditioned on a successor label.
That is, we let
$\Xi_{i, j}$
be the probability that a node has label $j$,
given that one of its successors has label $i$.

Formally, 
\begin{equation}\label{parametro_Theta}
    \Theta_{i, j} =  
    \mathbb{P} (Y_v = j \; | \;
    \exists u \in N^\text{in}(v) \; \text{ s.t. } Y_u = i),
\end{equation}

\begin{equation}\label{parametro_Xi}
    \Xi_{i, j} =  
    \mathbb{P} (Y_v = j \; | \;
    \exists u \in N^\text{out}(v) \; \text{ s.t. } Y_u = i),
\end{equation}

for all $i, j \in \mathcal{Y}$,
also note that
$\sum_{j=1}^K \Theta_{i,j}
= \sum_{j=1}^K \Xi_{i,j} = 1$ for all $i \in \mathcal{Y}$.

Denote by
$D_v^{\text{in}}$
and 
$D_v^{\text{out}}$
the random variables indicating
the in and out degree of $v$.
For each $i \in \mathcal{Y}$,
let $\psi_i$ (and $\phi_i$) 
denote the probability mass functions
of $D_v^{\text{in}}$
(and $D_v^{\text{out}}$),
conditioned on the event $Y_v = i$.
Namely,

\begin{equation}\label{parametro_psi}
    \psi_i (d) = f_{D_v^{\text{in}} |  Y_v} (d, i)
    = \mathbb{P} (D_v^{\text{in}} = d \,\, |\,\, Y_v = i),
\end{equation}

\begin{equation}\label{parametro_phi}
    \phi_i (d) = f_{D_v^{\text{out}} | Y_v} (d, i)
    = \mathbb{P} (D_v^{\text{out}} = d \,\,|\,\, Y_v = i).
\end{equation}

Suppose that 
$D_v^{\text{in}} = d$
and $Y_v = i$,
where $d, i \in \mathbb{N}$ 
are known values.
Then $P_v$ is a random vector
of fixed sum
indicating the labels 
of the $d$ predecessors of $v$.
We assume that, conditioned on $Y_v = i$,
the labels of the predecessors are 
independent of each other
and follow the same distribution,
determined by the $i$-th row of $\Xi$,
denoted by $\xi_i \in \mathbb{R}^K$.
Consequently, given these conditions,
$P_v$ follows a multinomial distribution
with parameters $d$ and $\xi_i$, denoted by
$\{P_v | D_v^{\text{in}} = d, Y_v = i\}
\sim \text{Multinomial}(d, \xi_i)$.
Similarly, we get that
$\{S_v | D_v^{\text{out}} = d, Y_v = i\}
\sim \text{Multinomial}(d, \theta_i)$,
where 
$\theta_i \in \mathbb{R}^K$
is the $i$-th row of $\Theta$.
Additionally, note that
$\{P_v | D_v^{\text{in}} = 0, Y_v = i\}$
and 
$\{S_v | D_v^{\text{out}} = 0, Y_v = i\}$
follow the same distribution,
as both must be the zero vector 
of length $K$ in this case.

Finally,
for each $i \in \mathcal{Y}$,
let $\omega_i$ denote the 
probability mass function of $X_v$
conditioned on the event $Y_v = i$.
Namely,

\begin{align}\label{parametro_omega}
    \omega_i (x) &= f_{X_v | Y_v} (x, i)
    = \mathbb{P} (X_v = x\,\, |\,\, Y_v = i).
\end{align}

\section{Parameter estimation} \label{parameter_estimation}

Given a simple directed graph $G = (V, E)$,
where each node $v$
has associated attributes $x_v$ and labels $y_v$,
we assume the graph is governed 
by the probabilistic model 
described in Section \ref{modelo}.
Our goal is to estimate the parameters of this model.
Specifically,
we aim to estimate 
the vector $\pi$, 
the matrices $\Theta$ and $\Xi$,
as well as the functions $\psi_i$, $\phi_i$, and $\omega_i$ 
for each $i \in \mathcal{Y}$.

The parameters 
$\pi$, $\Theta$, and $\Xi$
can be estimated based on 
the observed data frequencies.
These estimators correspond to 
the maximum likelihood estimators
and provide unbiased estimates.
Nevertheless,
we can utilize additive smoothing
to avoid having zero probabilities.
This is useful to avoid
making estimations
of events with 
zero probability
and will be useful
when doing a 
node classification task with the model.
Specifically, let
$\alpha_{\pi},
\alpha_{\Theta},
\alpha_{\Xi} \geq 0$ 
represent the smoothing hyperparameters for
$\pi$, $\Theta$, and $\Xi$ respectively.
The case $\alpha = 0$
corresponds to frequency-based estimation
(i.e., no smoothing),
while $\alpha = 1$ applies Laplace smoothing.
The smoothed estimates 
$\hat{\pi}$, $\hat{\Theta}$, and $\hat{\Xi}$
are then computed as follows:

\begin{align*}
    \hat{\pi}_i &= 
    \frac{|\{v \in V: y_v = i\}| + \alpha_{\pi}}
    {n + K\alpha_{\pi}},
    \\
    \hat{\Theta}_{i, j} &= 
    \frac{|\{ (u, v) \in E: y_u = i, y_v = j \}| + \alpha_{\Theta}}
    {|\{ (u, v) \in E: y_u = i\}| + K\alpha_{\Theta}},  
    \\
    \hat{\Xi}_{i, j} &= 
    \frac
    {|\{ (v, u) \in E: y_v = j, y_u = i \}| + \alpha_{\Xi}}
    {|\{ (v, u) \in E: y_u = i\}| + K\alpha_{\Xi}}.
\end{align*}

We consider two main approaches 
to estimate the functions $\psi_i$ and $\phi_i$ 
for each label $i \in \mathcal{Y}$.
The first approach involves 
estimating these conditional probability functions
using additive smoothing,
with  hyperparameters 
$\alpha_{\psi}, \alpha_{\phi} \geq 0$.
Specifically, we select 
finite sets $D_{\psi}, D_{\phi} \subset \mathbb{N}_0$,
which define the degrees for which we want
to ensure positive estimates 
of $\psi_i$ and $\phi_i$, respectively.
That is, we assume 
$\psi_i(d) = 0$ for 
$d \in \mathbb{N}_0 \setminus D_{\psi}$
and $\phi_i(d') = 0$ 
for $d' \in \mathbb{N}_0 \setminus D_{\phi}$.
For 
$d \in D_{\psi}$
and $d' \in D_{\phi}$,
the estimates are given by:

\begin{align} \label{estimar_psi}
    \hat{\psi_i}(d)
    = \hat{\mathbb{P}}
    ( D_v^{\text{in}} = d| y_v = i)
    = 
    \frac{|\{ v \in V: d_v^{\text{in}} = d, y_v = i \}| + \alpha_{\psi}}
    { |\{ v \in V: y_v = i \}|
    + |D_{\psi}| \alpha_{\psi}},
\end{align}

\begin{align}  \label{estimar_phi}
    \hat{\phi_i}(d')
    = \hat{\mathbb{P}}
    ( D_v^{\text{out}} = d'| y_v = i)
    = 
    \frac{|\{ v \in V: d_v^{\text{out}} = d', y_v = i \}| + \alpha_{\phi}}
    { |\{ v \in V: y_v = i \}|
    + |D_{\phi}| \alpha_{\phi}}.
\end{align}

Alternatively,
since the random variables
$\{ D_v^{\text{in}} | Y_v = i\}$ 
and $\{ D_v^{\text{out}} | Y_v = i\}$
can take values over all of $\mathbb{N}_0$,
we can opt to fit a parametric distribution.
This approach reduces 
the number of parameters to estimate
and can provide greater interpretability to the model.
However, the choice of parametric family
depends on the specific case,
and its validity should be confirmed 
using a goodness-of-fit test.

The estimation of the functions $\omega_i$
is highly case-specific,
as the structure of the attribute set $\mathcal{X}$
can vary significantly across different applications.
We provide an example of estimating
the functions $\omega_i$ 
when each $x \in \mathcal{X}$
represents a document,
which is defined as a piece of text.
This is common in many real-world applications.
First, we define 
a vocabulary 
$\Sigma = \{ t_1, t_2, \dots, t_{\tau}\}$ 
consisting of $\tau$ unique terms,
and let $M$ be
the maximum length
that a document can have.
Then, the attribute set $\mathcal{X}$
consists of all possible sequences of at most 
$M$ terms from the vocabulary.
Formally, 
$
\mathcal{X} = \bigcup_{m = 1}^{M} \Sigma^{m}.
$
For each $x \in \mathcal{X}$,
the vector $\bar{x} \in (\mathbb{N}_0)^{\tau}$
is used to represent 
the frequency of each term $t_j$ in the document $x$.
The process of transforming the document $x$
into a vector $\bar{x}$ is referred to
as text vectorization.
In this context,
$\bar{x}_j$ represents
the number of occurrences of the term $t_j$ in $x$,
given by

\begin{equation}\label{vectorizar_texto}
    \bar{x}_j = 
    \sum_{t \in x} \mathds{1}_{(t = t_j)},
\end{equation}

for each $j \in \{1, \dots, \tau\}$.
Then, 
the estimated conditional probability 
$\hat{\omega}_i(x)$ 
of observing the text $x$
given the label $i$ is

\begin{equation}\label{estimar_omega}
    \hat{\omega}_i (x)
    = \hat{\mathbb{P}} (X_v = x | Y_v = i)
    = \frac{1}{M} \prod_{j=i}^{\tau} 
    (\hat{\eta}_{i, j}) ^ {\bar{x}_j}.
\end{equation}

Here, 
$\hat{\eta}_{i,j}$ 
is the estimated conditional probability
that the term $t_j$ appears in
the attributes of a node $v$
given $y_v = i$,
for each $i \in \mathcal{Y}$ 
and $j \in \{1, \dots, \tau\}$.
These probabilities are estimated
using additive smoothing with a hyperparameter
$\alpha_{\omega} \geq 0$
on the term frequencies across different classes,
namely

\begin{equation}\label{estimar_eta}
    \hat{\eta}_{i, j} = 
    \frac
    {\sum_{v \in V} \mathds{1}_{(y_v = i)} 
    (\bar{x}_v)_j  
    + \alpha_{\omega}
    }
    {\sum_{v \in V} 
    \mathds{1}_{(y_v = i)}
    \sum_{j = 1}^{\tau} (\bar{x}_v)_j  
    + \tau \alpha_{\omega}}.
\end{equation}

Note that
the parameters
$\hat{\eta}_{i,j}$
correspond to the conditional probabilities 
$\hat{\mathbb{P}}(t_j | y_v = i)$,
and their estimation,
as given by Equation \ref{estimar_eta},
follows a fundamentally similar procedure 
as the multinomial Naive Bayes classifier 
for text classification 
(see Equation \ref{eq_naive_bayes_proba_palabra}).
In fact, the example presented in this section 
for modeling the functions $\omega_i$
for text-based attributes is based on 
the probabilistic framework of multinomial Naive Bayes.
If we restrict our model 
to a graph with no connections between nodes,
the probabilistic inference performed by our model
for text attributes reduces
to multinomial Naive Bayes
applied to the nodes' text attributes.
In this sense, our probabilistic model
can be viewed as an extension of the Naive Bayes algorithm
to graph-structured data,
allowing it to incorporate both the content 
of the node and the graph structure for classification tasks.

\section{Node Classification}


Suppose we have a simple directed 
graph $G = (V, E)$,
with node attributes $x_v$ for each node $v$.
However, the labels $y_v$
are not available for all nodes.
Instead,
there is a subset of the nodes
$L \subset V$ 
for which 
$y_v$ is known for every $v \in L$,
while labels of nodes in $U := V - L$
are unknown. 
In that case,
the objective is to learn
a function $\hat{y}: U \rightarrow \mathcal{Y}$
such that $\hat{y} (v)$ is 
the label prediction for the node $v \in U$.
This task is referred to as node classification.

To predict the label of nodes $v \in U$,
we can assume that the graph follows 
the probabilistic model described in 
Section \ref{modelo}.
We first estimate the model parameters,
as outlined in Section \ref{parameter_estimation},
and then use these parameters 
to make label predictions.
In this section, we explain how to
predict node labels using the probabilistic model.

\subsection{Prediction over a single node} \label{node_class_single_node}

Consider a node $v \in V$ 
for which $y_u$ is known
for each $u \in N(v)$.
The prediction $\hat{y}(v)$
is derived from estimating the random variable $Y_v$.
Due to the assumption
that $Y_v$ is only dependent
on $P_v$, $S_v$ and $X_v$,
which are known,
we only use these values
for computing
$\hat{y}(v)$.
To simplify notation,
we drop the subscripts from the random variables,
as we are currently considering
only the information pertaining to a node $v$.
There are two
primary approaches to estimate $Y$:
maximum likelihood (ML) estimation 
and
maximum a posteriori (MAP) estimation.
We explain both methodologies.

\subsubsection{Maximum Likelihood Estimate}

The Maximum Likelihood (ML)
estimate of $Y$,
denoted by
$\hat{y}_{\text{MLE}}$,
is the solution to 
the following maximization problem.

\begin{equation} \label{MLE_principal}
    \hat{y}_{\text{MLE}} =
    \arg\max_{i \in \mathcal{Y}} 
    f_{P, S, X | Y} (p, s, x, i).
\end{equation}

Transforming this maximization problem
to a minimization problem and using 
of equation \ref{decomposition}, we obtain:

\begin{align} \label{mle_separada}
    \hat{y}_{\text{MLE}} =
    \arg\min_{i \in \mathcal{Y}}  -
     f_{P|Y}(p, i)
      f_{S|Y}(s, i)
      f_{X|Y}(x, i).
\end{align}

Since the random vector
$\{ P | D^{\text{in}} = d, Y = i\}$
follows a multinomial distribution 
with parameters $d$ and $\xi_i$, we can express:

\begin{align} \label{descomponer_p_dado_y}
    f_{P|Y} (p, i) 
    &= \sum_{d=0}^{\infty}
    f_{P, D^{\text{in}} | Y} (p, d, i) \notag  \\
    &= f_{P, D^{\text{in}} | Y} 
    \left(p, \sum_{k=1}^K p_k, i \right) \notag  \\
    &= f_{P, D^{\text{in}} | Y} 
    (p, d^{\text{in}}, i) \notag  \\
    &= f_{P | D^{\text{in}}, Y}
    (p, d^{\text{in}}, i )
    f_{D^{\text{in}}| Y}( d^{\text{in}}, i ) \notag \\
    &=  g(p; d^{\text{in}}, \xi_i ) 
    \psi_i( d^{\text{in}} ).
\end{align}

Using a similar argument, we get that

\begin{equation} \label{descomponer_s_dado_y}
    f_{S|Y} (s, i) = 
    g(s; d^{\text{out}}, \theta_i) 
    \phi_i(d^{\text{out}}).
\end{equation}

Substituting Equations 
\ref{descomponer_p_dado_y} and \ref{descomponer_s_dado_y},
plugging the estimations of the unknown parameters,
and taking into account that the logarithmic function
is monotonically increasing,
we rewrite Equation \ref{mle_separada} as

\begin{align*} \label{mle_final}
    \hat{y}_{\text{MLE}} =  
    \arg\min_{i \in \mathcal{Y}}  - 
    \text{log}(
    g(p; d^{\text{in}}, \hat{\xi}_i)
    ) -
    \text{log}(
    g(s; d^{\text{out}}, \hat{\theta}_i) 
    ) -
    \text{log}(
    \hat{\psi}_i(d^{\text{in}})
    )\\
     - \text{log}(
    \hat{\phi}_i(d^{\text{out}})
    ) - 
    \text{log}(
    \hat{\omega}_i(x)
    ).
\end{align*}

Since the set of labels
$\mathcal{Y}$ is typically not large,
the minimization can be efficiently performed
by evaluating all possible labels.

\subsubsection{Maximum A Posteriori Estimation}

The Maximum A Posteriori (MAP) estimation of $Y$,
denoted by
$\hat{y}_{\text{MAP}}$ is the solution to 
the following maximization problem:

\begin{equation*} 
    \hat{y}_{\text{MAP}} =
    \arg\max_{i \in \mathcal{Y}} 
    f_{Y | P, S, X} (i, p, s, x).
\end{equation*}

By applying Bayes's Theorem, 
we get that:

\begin{align*}
    \hat{y}_{\text{MAP}} &=
    \arg\max_{i \in \mathcal{Y}}     
    \frac{f_{P, S, X | Y} (p, s, x, i) f_{Y}(i)}
    {f_{P, S, X} (p, s, x)} \\
    &= \arg\max_{i \in \mathcal{Y}} 
    f_{P, S, X | Y} (p, s, x, i) f_{Y}(i)
\end{align*}

Note that the first term 
in this equation corresponds
exactly to the quantity 
in the minimization problem presented in Equation \ref{MLE_principal}.
Thus, we can derive

\begin{align*} \label{map_final}
    \hat{y}_{\text{MAP}} =  
    \arg\min_{i \in \mathcal{Y}}  -
    \text{log}(
    g(p; d^{\text{in}}, \hat{\xi}_i)
    ) -
    \text{log}(
    g(s; d^{\text{out}}, \hat{\theta}_i) 
    ) -
    \text{log}(
    \hat{\psi}_i(d^{\text{in}})
    )  \\
    - \text{log}(
    \hat{\phi}_i(d^{\text{out}})
    ) -
    \text{log}(
    \hat{\omega}_i(x)
    ) -
    \text{log}(
    \hat{\pi_i}
    ).
\end{align*}

Note that the MAP and ML estimates 
of $Y$ involve solving
almost the same minimization problem,
with the only difference being that the 
MAP estimate incorporates an additional term
corresponding to the prior distribution of $Y$.
When uniform priors are assumed in the MAP estimate,
i.e when
$\hat{\pi_i}$ is constant for all $i$,
it is equivalent to the ML estimate.
Additionally,
note that estimating labels
using this approach 
corresponds to a generative 
classifier.

\subsubsection{Interpretability}\label{section_interpretabilidad}

Both the ML and MAP estimates
are derived from minimization problems
that involve a sum of positive terms.
For each $i \in \mathcal{Y}$,
these positive terms represent discrepancies
between specific information in the graph
and the belief that the true label is $Y = i$.
Then, the minimization process selects
the label that minimizes these discrepancies, 
thereby reflecting the label that is
most congruent with the observed data.

Each term in the minimization problem provides
insight into how well a potential label
aligns with different aspects of the graph.
A term with a high value indicates
a significant discrepancy
between the label and a particular piece of information.
The interpretation of each term is as follows:

\begin{itemize}
    \item Label Predecessors Discrepancy:
    $-\text{log}(g(p; d^{\text{in}}, \hat{\xi}_i))$.
    
    \item Label Successors Discrepancy:
    $-\text{log}(g(s; d^{\text{out}}, \hat{\theta}_i))$.
    
    \item Predecessor Count Discrepancy:
    $-\text{log}(\hat{\psi}_i(d^{\text{in}}))$.
    
    \item Successor Count Discrepancy:
    $-\text{log}(\hat{\phi}_i(d^{\text{out}}))$.
    
    \item Attribute Discrepancy:
    $-\text{log}(\hat{\omega}_i(x))$.
    
    \item Prior Belief Discrepancy:
    $-\text{log}(\hat{\pi_i})$.
\end{itemize}

By utilizing the probabilistic model for prediction,
we can identify the elements of the node
that support the assigned label.
Furthermore, we can analyze which factors
might have led to the exclusion of alternative labels,
thereby enhancing our understanding 
of the underlying mechanisms 
influencing the model's predictions.
In Section \ref{ejemplo_prediccion}
we provide examples to illustrate
the interpretability of the predictions
computed by the model.

\subsection{Prediction over several nodes}

In many cases, 
the objective is to predict the label of a node
$v \in V$
for which one or more of its neighboring 
nodes have unknown labels,
i.e $\exists u \in N(v)$
such that $u \in U$.
Under these circumstances,
the method described in Section
\ref{node_class_single_node}
cannot be directly applied to compute the prediction
$\hat{y}(v)$.
Instead, we can first compute an estimate
$\hat{y}'(u)$
for the neighboring node $u$
and use this estimate
as a substitute for the actual value of $y_u$
in order to predict $y_v$
using the previously outlined methodology.
While the estimate
$\hat{y}'(u)$
may not be completely precise,
it is unlikely to have a significant impact
on $\hat{y}(v)$, 
as this estimate depends on multiple factors.

To compute $\hat{y}'(u)$,
any straightforward heuristic can be employed.
We propose to utilize a simplified version 
of the method described in Section \ref{node_class_single_node},
where the information of the neighborhood
of $u$ is not utilized.
Consequently, the estimation is given by

\begin{equation}\label{iteracion_0_texto}
    \hat{y}'(u) =
    \arg\min_{i \in \mathcal{Y}} 
    - \text{log}(
    \hat{\omega}_i(x_u)).
\end{equation}

Moreover, we explore another heuristic
for computing $\hat{y}'(u)$,
which is by examining near nodes.
Specifically,
the neighborhoods of different orders of $u$ 
are explored
until a node with known label $w$ is found,
then $\hat{y}'(u) = y_w$.
If no node has known label
in the explored neighborhoods,
then $\hat{y}'(u)$ is randomly selected.
For the experiments,
we consider both approaches for computing
the simple estimations $\hat{y}'(u)$.
We select the approach that gives the 
best results,
treating this option as a hyperparameter.

Generalizing this idea to the whole graph,
we learn the function 
$\hat{y}: U \rightarrow \mathcal{Y}$
through an iterative process.
In this framework, 
we utilize simple estimates of all unknown 
labels to generate improved estimates,
iteratively refining these predictions.
This process is described in 
Algorithm \ref{algo_predictions},
where the estimations can be derived 
using either ML or MAP estimates.
This iterative process is advantageous,
as it can be efficiently parallelized,
thereby significantly reducing 
computational time and making 
it well-suited for large datasets.
The iterative process of 
generating estimations 
can be stopped
after a fixed number of iterations,
or whenever the 
label predictions
remains constant for most of the nodes.
The outcome of this process is a 
set of estimations
$\{ \hat{y}_{v}^{(0)}, \hat{y}_{v}^{(1)}, 
\dots, \hat{y}_{v}^{(T)} \}$ 
for each node $v \in U$.
To select the final estimations $\hat{y}$,
a specific iteration $t$ 
must be chosen, leading to the assignment
$\hat{y}(v) = \hat{y}_{v}^{(t)}$,
for all $v \in V$.
The selection
of iteration $t$
can be achieved 
by defining a validation set of nodes
with known labels,
and subsequently selecting 
the iteration that yields the
best performance on this validation set.

\begin{algorithm}[H]
    \caption{Prediction over several nodes}
    \label{algo_predictions}
    \begin{algorithmic}
    \footnotesize

    \STATE \textbf{Input:} 
    Attributed directed graph $G = (V, E)$
    with labels defined for a set 
    $L \subset V$ (labeled nodes),
    number of iterations $T$, 
    and convergence tolerance $\epsilon \in (0, 1)$.

    \STATE \textbf{Output:} 
    Estimations
    $\{ \hat{y}_{v}^{(0)}, \hat{y}_{v}^{(1)}, 
\dots, \hat{y}_{v}^{(T)} \}$
    for each node $v \in U = V \setminus L$ (unlabeled nodes).

    \STATE 
    $\hat{y}_{v}^{(0)} \leftarrow$
    estimation $\hat{y}'(v)$
    using any simple method,
    for each $v \in U$.

    \STATE $t \leftarrow 1$

    \REPEAT
        \FORALL {$v \in U$}
            \STATE Update $\hat{y}_{v}^{(t)} 
            \leftarrow \hat{y_v}$ assuming 
            $\hat{y}_{v}^{(t-1)}$ as the real labels
            of the neighbors.
        \ENDFOR

        \STATE $t \leftarrow t + 1$

    \UNTIL{$\hat{y}_{v}^{(t)} = \hat{y}_{v}^{(t-1)}$
    for at least $100(1 - \epsilon)\%$ of nodes $v \in U$, or
    $ t > T$}

    \end{algorithmic}
\end{algorithm}

\section{Math Genealogy Project}


The Math Genealogy Project 
is an initiative aimed at 
compiling comprehensive information 
about mathematicians worldwide\footnote{available online at: \url{https://mathgenealogy.org/index.php}}.
The goal is to collect data about individuals
who have obtained 
doctorates in mathematics, statistics,
computer science, mathematics education, 
operations research or similar.
This dataset is significant for understanding
the lineage and contributions of mathematicians,
which can illuminate 
the evolution of mathematical thought.
For each mathematician, 
the math genealogy project
records:

\begin{itemize}
    \item The full name of the individual.
    \item The year the degree was awarded and the university granting it.
    \item The complete name(s) of the advisor(s).
    \item The title of the dissertation.
    \item The Mathematics Subject Classification (MSC)
    for the dissertation.
\end{itemize}

The MSC is a  
classification scheme used
by numerous mathematics journals
to categorize research topics.
The MSC is hierarchical,
encompassing 63 mathematical disciplines
at its highest level. 
This classification is 
particularly useful
in the Math Genealogy Project
for identifying a mathematician's primary
area of interest and research focus.

\subsection{Learning task}

The dataset from the Math Genealogy Project is incomplete,
with several mathematicians 
lacking certain fields of information.
Consequently, it is of interest 
to predict these missing values using the available data. 
Specifically, we aim to 
predict the Mathematics Subject Classification (MSC)
of each mathematician.
To accomplish this,
we train classification algorithms
that leverage the advisor-student relationships, 
as well as the titles of dissertations
associated with each mathematician, to predict the MSC.
The first step in making predictions
was to acquire the dataset.
The web page of the Math Genealogy Project
was fetched to gather all available information.
This data collection procedure
was conducted on October 2, 2023,
using a modified version of 
publicly available code for
scraping the math genealogy website\footnote{GitHub repository: \url{https://github.com/j2kun/math-genealogy-scraper/tree/main}}.

\begin{figure}[ht]
    \centering
    \includegraphics[width=0.7\linewidth]{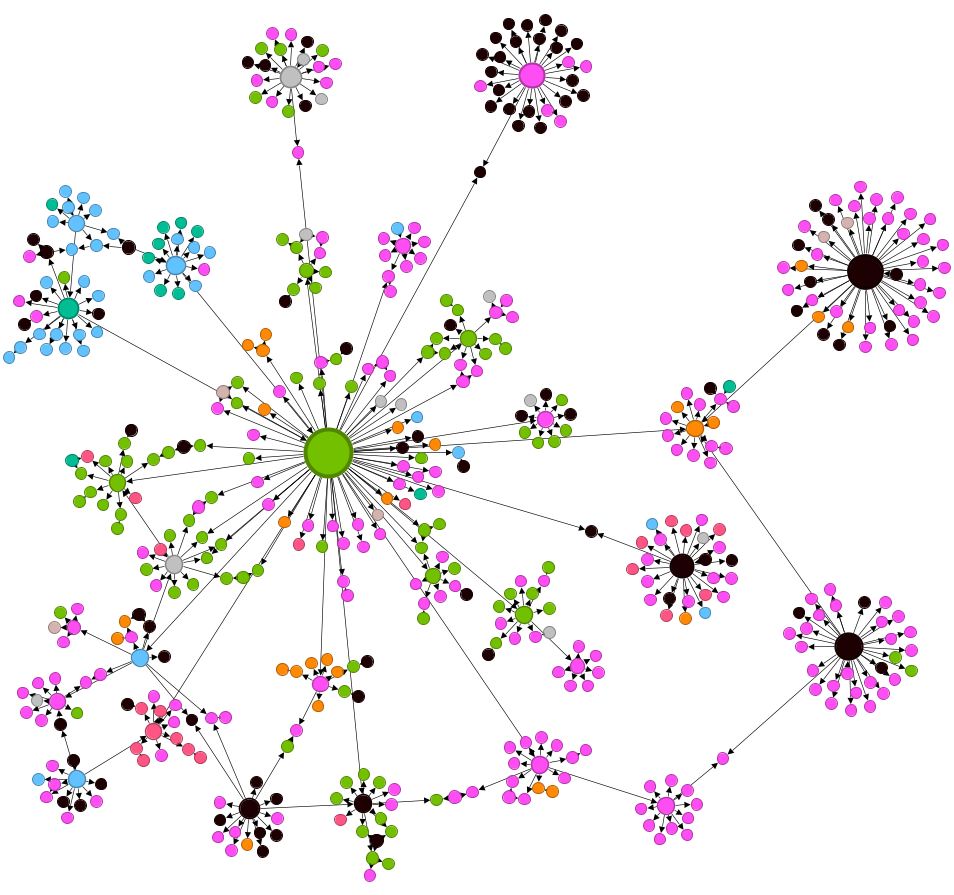}
    \caption{Induced subgraph of size 500
    of the resulted directed graph 
    representing the information
    of the Math Genealogy Project.
    The size of a node is given by its out degree.
    The distinct colors represent different
    MSC of the nodes, where the nodes with missing
    MSC are represented in black.}
    \label{fig:mgp}
\end{figure}

Based on the fetched dataset,
a directed graph was constructed
where mathematicians serve as nodes,
with directed edges going
from each advisor to their respective students.
The resulting directed graph comprises
297,377 nodes and 329,209 edges. 
We utilize 
the titles of dissertations 
as node attributes.
Therefore,
we filtered the graph to include
only those nodes with available dissertation titles,
resulting in a reduced graph 
with 267,774 nodes and 281,288 edges.
Notably, over 90\% of the mathematicians 
in the dataset have their dissertation titles available,
affirming the validity of this attribute for our predictions.
This resulting graph is the one
considered for the node
classification task.
In figure \ref{fig:mgp}
we present a visualization
of an induced subgraph,
where colors with different
MSC classification are drawn
using distinct colors.

Out of the 267,774 mathematicians
with available dissertation titles,
only 174,501 
possess corresponding MSC classifications,
which accounts for approximately 65\% of the graph.
The 93,273 nodes 
with dissertation titles but lacking MSC labels
are retained to enhance graph connectivity
and facilitate information flow within the network.
The subset of 174,501 nodes
with both dissertation titles and MSC classifications
will be employed for 
training and testing the different classification algorithms.

Following standard machine learning protocols,
we partition this subset into three categories:
the training set, containing 141,345 nodes,
is used to estimate the parameters
of the models;
the validation set, consisting of 15,705 nodes,
is employed to select the
optimal hyperparameters;
and the testing set, comprising 17,451 nodes,
is utilized to evaluate 
the predictive performance of the models.
Thus, the final graph comprises
52.79\% training nodes, 
5.87\% validation nodes,
6.52\% testing nodes,
and 34.83\% nodes without MSC classifications.
This is the division considered
in the following sections.
Additionally,
the same evaluation metric 
is employed to assess the performance 
of the various algorithms.
Due to the class imbalance in the MSC labels,
we propose using the F1-score 
as an appropriate evaluation measure.

\subsection{Application of the model}\label{model_in_mgp}

The main
classification algorithm
employed for the Math Genealogy Project
is our probabilistic model.
As outlined in Section \ref{parameter_estimation},
estimating the parameters of the model
necessitates an adaptation to 
the specific characteristics of the data.
In this section, we detail 
the process of adjusting the model to accommodate
the Math Genealogy Project graph,
where the titles of dissertations 
serve as node attributes and the 
Mathematics Subject Classification (MSC)
represents the node labels.
To begin, note that 
the key parameters of the graph
are given by:
$n = 267,774$, the number of mathematicians;
$m = 281,288$, the number of advisor-student relationships;
and $K = 63$, the total number of unique MSC categories.

\subsubsection{Estimating the functions $\omega_i$} \label{estimacion_omega_mgp}

The titles of dissertations
serve as node attributes, 
meaning that 
for every node $v$,
the attribute
$x_v \in \mathcal{X}$
is a piece of text.
Therefore, 
to estimate the conditional probabilities
$\omega_i$ we follow the procedure outlined
in Section \ref{parameter_estimation}.
To construct the vocabulary $\Sigma$,
we include all unigrams
and bigrams
(i.e. groups of one or two words)
that appear at least twice in the corpus,
which consists of the titles of dissertations
from the training nodes,
with English stop words removed.

The vectors $\bar{x}_v$
are obtained using Equation
\ref{vectorizar_texto},
and
the parameters $\eta_{i,j}$
are estimated through 
additive smoothing with a hyperparameter
$\alpha_{\omega} > 0$,
as described in Equation
\ref{estimar_eta}.
The smoothing hyperparameter 
$\alpha_{\omega}$ is optimized
using the validation set.
Finally, the estimation
of $\omega_i$ is carried out
according to Equation \ref{estimar_omega},
with the difference that
the term $\frac{1}{M}$
is omitted.
Note that this term is not relevant
for label estimation,
as it is a constant
within the optimization problem.
Consequently, 
the estimated functions 
$\hat{\omega}_i$ are proportional to,
but not equal to,
the conditional probabilities,
which suffices for the node classification task.
Specifically,

\begin{equation*}
    \sum_{x \in \mathcal{X}} \hat{\omega}_i(x) = C,
\end{equation*}

for all $i \in \mathcal{Y}$,
where $C$ is a constant.
Furthermore, we explored 
an alternative approach for text vectorization,
different than Equation \ref{vectorizar_texto}.
The vectors $\bar{x}_v$
can also be computed using a 
Term Frequency-Inverse Document Frequency (TF-IDF)
transformation 
\cite{tf_idf}.
This transformation generates 
vectors $\bar{x}_v$
that quantify the importance of
each term within the respective text,
resulting in entries
that can take any positive real number.
The parameters $\eta_{i,j}$
and the estimations $\hat{\omega}_i$
are computed using the same procedure,
following Equations \ref{estimar_eta}
and \ref{estimar_omega}
(without considering the term $\frac{1}{M}$).
Under this approach,
the estimated functions
$\hat{\omega}_i$
are proportional
to the conditional probabilities; 
however, the normalization 
term may vary across classes.
Namely,

\begin{equation*}
\sum_{x \in \mathcal{X}}
\hat{\omega}_i(x) = C_i    ,
\end{equation*}

where $C_i$ is a constant
dependent of the label $i \in \mathcal{Y}$.
Employing these functions to
perform node classification
requires the assumption
$C_i \approx C_j$
for all $i, j \in \mathcal{Y}$.
Both approaches of text vectorization
are considered in our experiments,
treating the selection of the methodology
as a hyperparameter chosen based
on the validation set results.

\subsubsection{Estimating the functions $\phi_i$}\label{estimacion_phi_mgp}

Note that
$\phi_i(d)$
represents
the probability that a node
with label $i$ has $d$ students,
where $d$ is a non-negative integer.
To model these conditional probabilities,
we propose probability mass functions parameterized by
$\beta_i \in (0, 1)$,
$\kappa_i > 0$ and $\lambda_i > 0$.
The parameter
$\beta_i$
directly defines the probability
that a mathematician with label $i$
has no students. 
Additionally, 
the proposed parametric functions
follow a truncated power law distribution
with parameters 
$\kappa_i$ and $\lambda_i$
when restricted to the positive integers.
Formally, we define the
probability mass functions as follows:

\begin{equation}\label{probabilidad_estudiantes}
\phi_i(d; \beta_i, \kappa_i, \lambda_i) =   
\begin{cases}
    \beta_i 
    & \text{ if } d = 0, \\
    (1 - \beta_i)
    \frac{d^{-\kappa_i} e^{-d\lambda_i}}
    {\sum_{d' = 1}^{\infty}
    d'^{-\kappa_i} e^{-d'\lambda_i}}
    & \text{ if } d \geq 1. \\
\end{cases}
\end{equation}

The estimation
$\hat{\beta_i}$
is computed as
the proportion of nodes with
label $i$ that do not have any students.
Then, the estimations
$\hat{\kappa_i}$ and $\hat{\lambda_i}$
are computed 
by maximum likelihood
via a computational program.
Consistent with the estimation of other parameters,
the estimations
$\hat{\beta_i}$,
$\hat{\kappa_i}$ and
$\hat{\lambda_i}$
are computed using
the samples of out-degrees
obtained
considering only the training nodes.
However, the out-degree of each node in the training 
set is computed by considering all nodes in the graph,
rather than restricting it to other training nodes.
This approach is similarly applied
to the estimation of the functions
$\psi_i$.

\begin{figure}[ht]
    \centering
    \includegraphics[width=0.99\linewidth]{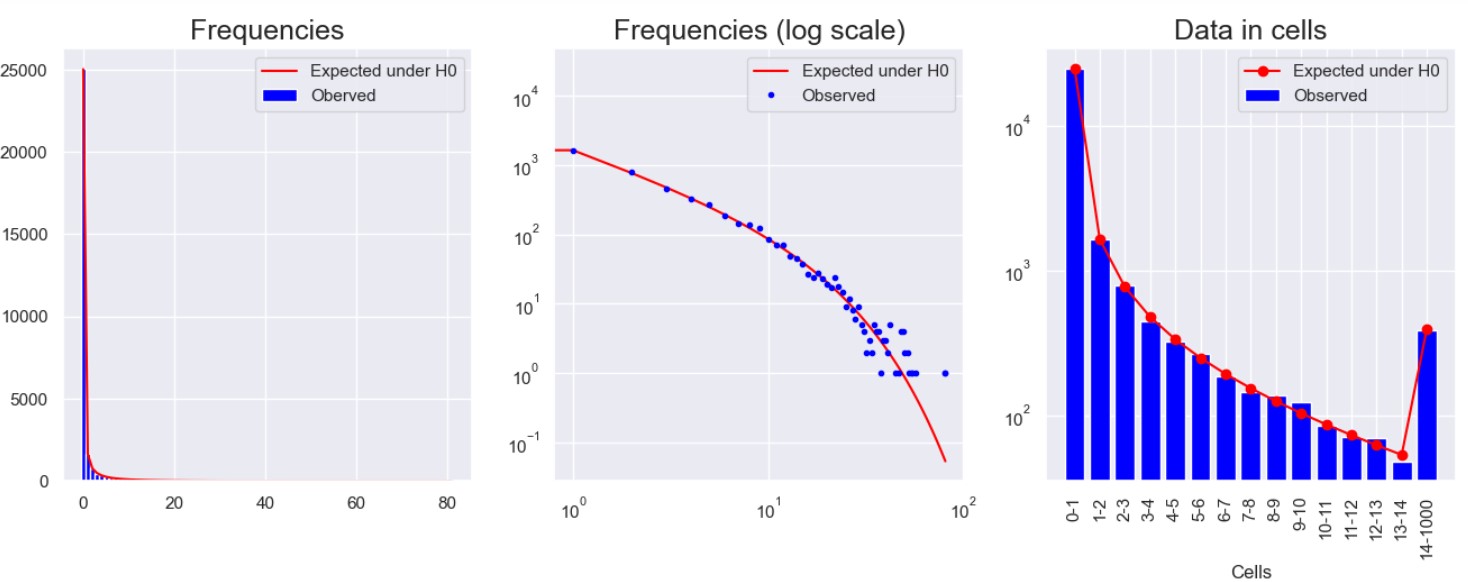}
    \caption{Chi-squared goodness-of-fit for the out degree of nodes with label 46 (68—Computer science).
    For this sample, the test statistic
    takes the value $T = 12.109$, 
    yielding a p-value of 0.35.
    Therefore, the test concludes that
    the data
    follows the proposed distribution.}
    \label{fig:mgp_out_46}
\end{figure}

To validate the suitability
of the parametric functions 
outlined in Equation \ref{probabilidad_estudiantes},
we employ goodness-of-fit hypothesis tests.
These tests incorporate 
all available information 
within the graph,
rather than just considering training nodes,
to enhance accuracy of the results.
For each label $i \in \mathcal{Y}$,
we conduct a Chi-squared goodness-of-fit
test with $k = 15$ cells.
If the expected cell values are not valid
(i.e.
if any value is below 1 or more than 20\%
of the values are below 5),
the number of cells is reduced
until the expected values meet the validity criteria.
Then, the p-value is calculated using a 
chi-square distribution with
$15 - 1 - 3 = 11$ degrees of freedom
(or different when the number of cells is not 15).
We set the significance level at $0.05$.
Consequently, we do not reject the null hypothesis 
that the data follow the proposed distribution
if the resulting p-value exceeds 
$0.05$; in such cases,
we conclude that the data 
aligns with the proposed distribution.
An example of this methodology
is presented in Figure \ref{fig:mgp_out_46},
were we show the goodness of test test
associated with the 
out degree of nodes with label 46.

\begin{figure}[ht]
    \centering
    \includegraphics[width=0.99\linewidth]{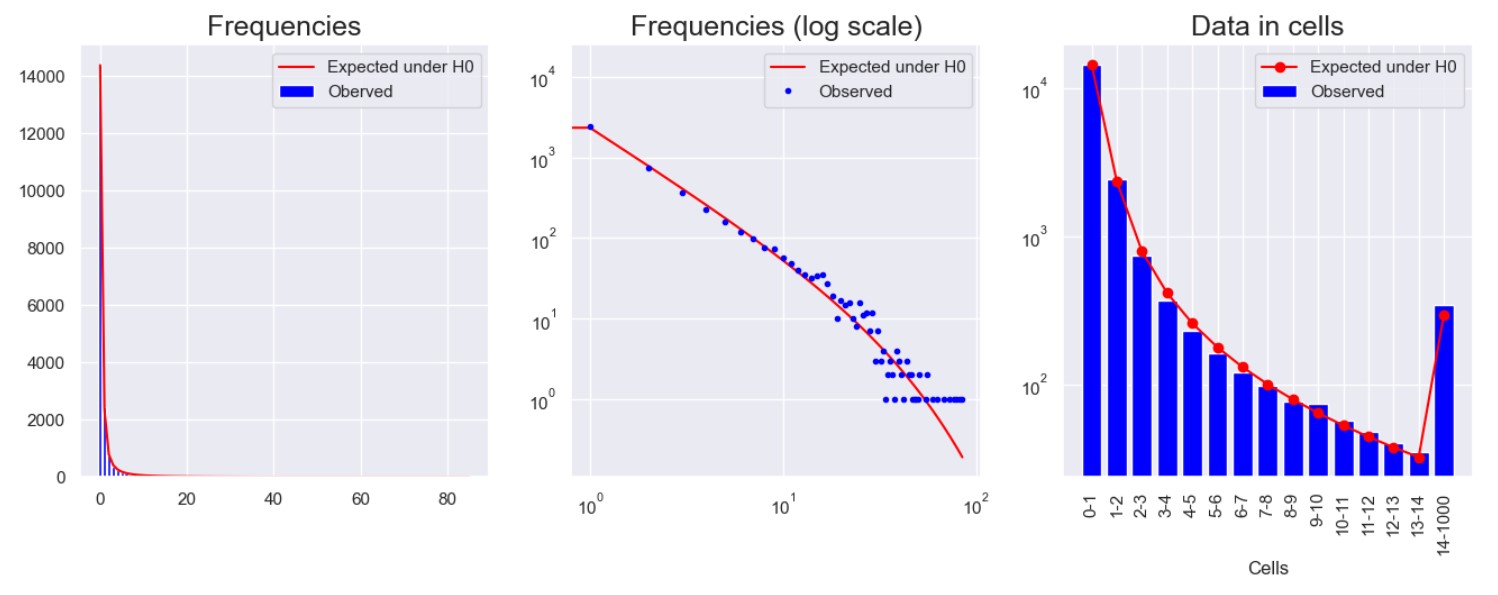}
    \caption{Chi-squared goodness-of-fit for the out degree of nodes with label 
    58 (91—Game theory, economics, social and behavioral sciences).
    For this sample, the test statistic
    takes the value $T = 25.5$, 
    yielding a p-value of 0.007.
    Therefore, the test concludes that
    the data does not
    follow the proposed distribution.
    However, this distribution appears to be a good approximation.}
    \label{fig:mgp_out_58}
\end{figure}

Following this methodology,
we find out that
58 out of the 63 samples
follow the proposed distribution,
this is a 92.06\%
success rate.
Moreover,
even in unsuccessful cases,
the proposed distribution
appears to be a good approximation
of the data distribution,
an example of an unsuccessful case
is presented in Figure
\ref{fig:mgp_out_58}.
Given that the majority of labels
appear to adhere to the proposed distribution,
we employ these parametric functions to model the functions 
$\phi_i$.

\subsubsection{Estimating other parameters}

To estimate the 
transition probabilities
$\Theta, \Xi$ and the
probabilities $\pi$,
we employ additive smoothing
as outlined in Section 
\ref{parameter_estimation}.
The hyperparameters 
$\alpha_{\pi},$
$\alpha_{\Theta}$, and
$\alpha_{\Xi}$ 
are optimized using the validation set.
Regarding the estimation of the functions $\psi_i$,
note that $\psi_i(d)$
represents the probability that a node
with label $i$ has $d$ advisors.
The number of advisors is inherently limited,
98.5$\%$ of mathematicians
have at most 2 advisors, 
with a maximum observed count of 6 advisors.
Consequently,
it is not necessary 
to fit a parametric distribution 
for estimating these functions.
Instead, we utilize 
additive smoothing with hyperparameter
$\alpha_{\psi}$,
which is also optimized using the validation set.

\subsection{Baselines for subject classification}

In addition to our proposed probabilistic model,
we employed other classification algorithms
to predict the Mathematics Subject Classification (MSC)
of mathematicians.
This approach enables us
to compare the performance 
of our model with established methods.
First,
we recognize two primary approaches 
to the MSC classification task:

\begin{itemize}
    \item Utilizing only the titles of dissertations, treating it as a text classification task.
    \item Incorporating both the titles of dissertations and graph information, treating it as a node classification task.
\end{itemize}

Furthermore, we categorize the potential
classification methods into two main types:

\begin{itemize}
    \item Models based on probabilistic inference.
    \item Models based on neural networks.
\end{itemize}

To encompass a 
diverse array of classification algorithms,
we selected one algorithm 
for each combination of approach and method.
Our probabilistic model corresponds
to an algorithm based on probabilistic inference
for the node classification task.
For text classification,
we implemented a Naive Bayes algorithm, 
which serves as a model based on probabilistic inference.
In the context of neural network-based algorithms,
we opted for a 
Bidirectional Encoder Representations from Transformers (BERT)
model for text classification
and a Graph Convolutional Network (GCN) 
for node classification.
The categorization of algorithms 
utilized for MSC classification 
is summarized in Table \ref{tabla_clasificacion_algoritmos}.
Additionally,
we evaluated Label Propagation
\cite{label_propagation} on this dataset,
as this method has demonstrated superior performance 
compared to certain graph neural network 
models and 
serves as a common benchmark for node classification tasks.

\begin{table}[ht]
    \begingroup
    \renewcommand{\arraystretch}{0.8} 
    \centering
    \begin{tabular}{@{} l| c c @{}} 
    \toprule
    \textbf{Classification} & \textbf{Probabilistic Inference} & \textbf{Neural Networks} \\ 
    \midrule
    \textbf{Text Classification} & Naive Bayes & BERT \\ 
    \textbf{Node Classification} & Our Model & GCN \\ 
    \bottomrule
    \end{tabular}
    \caption{Classification of Algorithms Used for MSC Classification.}
    \label{tabla_clasificacion_algoritmos}
    \endgroup
\end{table}

\subsection{Results on the Math Genealogy Project Dataset}

This section presents the best hyperparameters selected
for each method,
and the classification results
on the test set.
Hyperparameters were tuned on the validation set based on F1-score.

\subsubsection{Hyperparameter Selection}

\textbf{Naive Bayes:} 
Given its low training time,
we explored a broad hyperparameter space via grid search.
The selected hyperparameters are as follows:

\begin{itemize} 
    \item \textbf{N-gram range:} Unigrams and bigrams (1-2).
    \item \textbf{Minimum term frequency threshold:} 1 (no threshold).
    \item \textbf{Maximum vocabulary size:} No limit (551,776 terms).
    \item \textbf{Additive smoothing parameter:} $\alpha = 0.01$.
    \item \textbf{Prior distribution:} Estimated from label frequencies.
\end{itemize}

\textbf{BERT:} 
Due to the computational intensity of fine-tuning,
we explored a smaller hyperparameter space,
selecting the best configuration as follows:

\begin{itemize}
    \item \textbf{Weight decay:} 0.01.
    \item \textbf{Loss function:} Non-weighted.
    \item \textbf{Epochs:} 7.
    \item \textbf{Learning rate:} $10^{-5}$.
\end{itemize}

\textbf{Graph Convolutional Network (GCN):} 
For GCN, 
the graph is treated as undirected,
considering both directions for every edge.
For the hyperparameters, we considered two variations in
the node feature matrix based on vocabulary size
(1500 or 108,672 features),
along with layer configurations, and loss function options
(weighted or non-weighted).
The best hyperparameters are:

\begin{itemize}
    \item \textbf{Feature size:} 
    $d^{(0)} = 108,672$ (all terms in the training set).
    \item \textbf{Loss function:} Non-weighted.
    \item \textbf{Layer configuration:} Two layers, $d^{(1)} = 300$, $d^{(2)} = 300$.
    \item \textbf{Learning rate:} 0.01.
    \item \textbf{Training epochs:} 100, with early stopping at 10 consecutive epochs without validation improvement.
\end{itemize}

\textbf{Proposed Probabilistic Model:}
The hyperparameters for the probabilistic model 
are described in Section \ref{model_in_mgp},
we consider MAP and ML estimations 
using the same parameters and hyperparameters.
We performed six iterations
of generating predictions,
selecting the final iteration as the one that achieved the best validation performance.
Table \ref{hyper_mgp} summarizes the selected hyperparameters.

\begin{table}[ht]
    \begingroup
    \renewcommand{\arraystretch}{0.8} 
    \centering
    \begin{tabular}{@{} l c @{}}
    \toprule
        \textbf{Hyperparameter} & \textbf{Selection} \\ 
    \midrule
        Text vectorization method & TF-IDF \\
        N-gram range & Unigrams and Bigrams (1-2)  \\  
        Minimum document frequency & 2 \\ 
        Maximum document frequency & 0.5 \\ 
        Max features in vocabulary & No limit \\ 
        $\alpha_{\omega}$ & 0.03 \\
        $\alpha_{\pi}$ & 0 \\ 
        $\alpha_{\Theta}$ & 1 \\ 
        $\alpha_{\Xi}$ & 1 \\ 
        $\alpha_{\psi}$ & 0.1 \\  
        Estimation at iteration 0 & Nearest node \\
        Best iteration ML &   5 \\
        Best iteration MAP &   5 \\
    \bottomrule
    \end{tabular}
    \caption{Selected Hyperparameters of our model for MSC Classification.}
    \label{hyper_mgp}
    \endgroup
\end{table}

\subsubsection{Prediction Results and Discussion}

In Table \ref{resultados_mgp} we present the F1-score and accuracy
for each classification method on the test set.
Naive Bayes shows the lowest predictive performance
among all methods,
with an F1-score of 0.4726.
This is expected, as Naive Bayes relies solely on textual data,
and it is known to be outperformed by more complex algorithms.
BERT, while a powerful model for text classification, 
achieves an F1-score of 0.5028,
performing slightly better than Naive Bayes
Despite being state-of-the-art in text classification, 
BERT is outperformed in this data,
highlighting the importance of graph information
in the Math Genealogy Project dataset.
The GCN model, which considers both node features and graph structure,
surpasses both text-only models with an F1-score of 0.5689.
This improvement demonstrates the benefit of using 
a model that integrates 
structural information of the graph.
Our probabilistic model shows
the strongest results overall,
with maximum likelihood (ML) estimations
that achieve the highest F1-score of 0.5704, 
slightly outperforming the GCN model.
On the other hand, the MAP estimation of our model achieve
the highest accuracy across all methods at 0.7463.
This divergence in performance metrics between MLE and MAP is consistent
with the expected characteristics of these estimations.
ML achieves better F1-score due 
to balanced predictions across classes,
while MAP is more biased towards frequent labels,
therefore improving on accuracy but not necessarily in F1-score.
In summary, our probabilistic model achieves competitive F1-scores
and the highest accuracy from the four considered methods,
demonstrating its effectiveness
for classification tasks.

\begin{table}[ht]
    \begingroup
    \renewcommand{\arraystretch}{0.9} 
    \centering
    \begin{tabular}{@{} l c c @{}}
    \toprule
        \textbf{Method} & \textbf{F1 score} & \textbf{Accuracy} \\ 
    \midrule
        Our Model (ML)   & \textbf{0.5705}  & 0.7362 \\ 
        Our Model (MAP)  & 0.5495           & \textbf{0.7463} \\ 
        GCN              & 0.5689           & 0.7076 \\ 
        BERT             & 0.5028           & 0.6695 \\ 
        Naive Bayes      & 0.4726           & 0.6562 \\ 
        Label Propagation& 0.4812           & 0.6436 \\ 
    \bottomrule
    \end{tabular}
    \caption{Results for Testing Data in MSC Classification.}
    \label{resultados_mgp}
    \endgroup
\end{table}

\subsection{Prediction example}\label{ejemplo_prediccion}

In this section,
we present examples of MSC predictions 
computed using our probabilistic model.
This highlights the interpretability 
our model provides, compared to neural networks 
and other machine learning models,
which generally lack 
such interpretative clarity. 
The examples presented correspond
to MAP estimations at iteration 5,
which achieved the best prediction
performance on the validation set.
When predicting the label of a node $v$,
if any neighbor $u \in N(v)$ 
lacks a known label (i.e., it is not in the training set),
its predicted label from iteration 4,
obtained using MAP,
is treated as its true label.
This process follows the steps outlined in Algorithm \ref{algo_predictions}
For each example,
we display the discrepancies between
the observed data and 
the three labels with the lowest discrepancies
(i.e. the most suitable labels), 
according to the method detailed in Section 
\ref{section_interpretabilidad}.

The first example considers the classification
of the node with ID 1526,
which is part of the test set,
meaning its true label is excluded during training.
The true label for this node is 41, 
representing the MSC class:
57—Manifolds and cell complexes.
The dissertation title for this node is: 
``On Topological Vector Fields''.
This node has two predecessors (advisors),
both of which are not training nodes.
Thus, the model uses their predicted labels
from the prior iteration:
label 40 for one predecessor 
and label 41 for the other. 
The node also has twelve successors (students),
five of whom are labeled training nodes 
with the true label 41. 
For the remaining seven students, 
the model uses the previous 
iteration’s predictions, which assign label 41 to each.
Based on this information,
we calculate the discrepancies
across each component of the data.

Table \ref{ejemplo_nodo_12408}
shows the discrepancy values across the 
three labels with the least total discrepancy:
label 41 (57—Manifolds and cell complexes),
label 40 (55—Algebraic topology),
and label 42 (58—Global analysis, analysis on manifolds). 
The model ultimately predicts label 41, 
as it minimizes the overall discrepancy.
The table also illustrates that,
if only the dissertation title 
(the attribute data) was considered, 
labels 40 or 42 would be preferred.
However, these labels are not selected 
due to their significantly higher 
successor label discrepancies, 
label 41 is a more coherent choice given 
the successors’ labels.
Thus, even though label 41 is not 
the closest fit based solely on the attribute, 
it is the predicted label due to 
its alignment with the broader label structure
of the node's successors.

\begin{table}[ht]
    \begingroup
    \renewcommand{\arraystretch}{0.9} 
    \centering
    \begin{tabular}{@{} l c c c @{}}
    \toprule
        \textbf{Label} & 
        \textbf{41} & \textbf{40} & \textbf{42} \\ 
    \midrule
        Attribute discrepancy:
        & 21.32  & 19.41 & 20.32 \\ 
        Predecessor count discrepancy:
        & 2.08  & 2.13 & 1.98 \\ 
        Successor count discrepancy:
        & 6.21  & 6.37 & 6.14 \\ 
        Label Predecessors discrepancy:
        & 2.32  & 2.69 & 6.01 \\ 
        Label Successors discrepancy:
        & 10.59  & 29.56 & 37.95 \\ 
        Prior discrepancy:
        & 4.87  & 4.84 & 5.53 \\ 
        \textbf{Total Discrepancy:}
        & \textbf{47.42}  & \textbf{65.03} & \textbf{77.95} \\ 
    \bottomrule
    \end{tabular}
    \caption{Example of MSC classification on node $v$ with ID 12408.
    Information of the node:
    $x_v = $ ``On Topological Vector Fields'',
    number of predecessors
    $d^{\text{in}}_v = 2$,
    labels of predecessors
    $p_v = e_{40} + e_{41}$,
    number of successors
    $d^{\text{out}}_v = 12$,
    labels of successors
    $s_v = 12 e_{41}$,
    true label $y_v = 41$.}
    \label{ejemplo_nodo_12408}
    \endgroup
\end{table}

The second example considers 
the classification of the node with ID 153133,
which also belongs to the test set.
This node’s true label is 38,
corresponding to the MSC class:
53—Differential geometry.
The dissertation title for this node is
``Geometric objects in differential geometry''.
Node 153133 has one predecessor (one advisor)
who is also not part of the training set.
In the prior iteration,
the model predicted label 38 for this predecessor.
Additionally,
this node has two successors (students), 
both of which are labeled training nodes 
with a known label of 36.
Using this information,
we compute the discrepancy in each data component.

Table \ref{ejemplo_nodo_153133} 
presents the discrepancy values across 
the three labels with the lowest total discrepancies:
label 38 (53—Differential geometry),
label 36 (51—Geometry), 
and label 42 (58—Global analysis, analysis on manifolds).
The model's prediction for this node is label 38,
as it minimizes the total discrepancy.
The table allows us to analyze the
model’s reasoning behind this prediction.
Despite the successors' labels suggesting 
a strong preference for label 36,
the node's text attribute 
and the predecessor label
align with label 38.
Additionally, the prior beliefs
reinforce label 38 as the most suitable prediction.
Notably, 
the discrepancy scores for this prediction 
are closely matched, 
unlike in the previous example where the chosen label 
had a distinctly lower total discrepancy than the alternatives.
This similarity indicates a lower 
confidence level for this prediction,
which may be valuable information 
for certain applications where prediction 
certainty is a consideration.

\begin{table}[ht]
    \begingroup
    \renewcommand{\arraystretch}{0.9} 
    \centering
    \begin{tabular}{@{} l c c c @{}}
    \toprule
        \textbf{Label} & 
        \textbf{38} & \textbf{36} & \textbf{42} \\ 
    \midrule
        Attribute discrepancy:
        & 16.77  & 18.79 & 21.12 \\ 
        Predecessor count discrepancy:
        & 0.27  & 0.47 & 0.34 \\ 
        Successor count discrepancy:
        & 3.51  & 3.47 & 3.53 \\ 
        Label Predecessors discrepancy:
        & 0.46  & 2.22 & 1.93 \\ 
        Label Successors discrepancy:
        & 6.38  & 2.80 & 7.44  \\ 
        Prior discrepancy:
        & 3.98  & 4.95 & 5.53 \\ 
        \textbf{Total Discrepancy:}
        & \textbf{31.39}  & \textbf{32.72} & \textbf{39.91} \\ 
    \bottomrule
    \end{tabular}
    \caption{Example of MSC classification on node $v$ with ID 153133.
    Information of the node:
    $x_v = $ ``Geometric objects in differential geometry'',
    number of predecessors
    $d^{\text{in}}_v = 1$,
    labels of predecessors
    $p_v = e_{38}$,
    number of successors
    $d^{\text{out}}_v = 2$,
    labels of successors
    $s_v = 2 e_{36}$,
    true label $y_v = 38$.}
    \label{ejemplo_nodo_153133}
    \endgroup
\end{table}

\section{Ogbn-arxiv dataset}


The second dataset considered
for the application of our model is 
the \textit{ogbn-arxiv},
a widely-used benchmark dataset
that is part of the Open Graph Benchmark (OGB)
collection \cite{ogbn_arxiv_paper}. 
The ogbn-arxiv dataset was designed
for node classification tasks
(also referred to as node property prediction).
It consists of a directed citation network
where the nodes represent computer science papers
from the arXiv repository,
and each directed edge indicates that
the source paper cites the target paper. 
The graph contains $n = 169,343$ nodes
and $m = 1,166,243$ edges.
For each node, the dataset provides
the title and abstract of the respective paper. Additionally, a 128-dimensional vector
is provided for each node, 
which represents the average
of the word embeddings for the title and abstract.
In our experiments, we utilize
the raw text of the nodes 
(titles and abstracts with stop words removed)
instead of the precomputed embedding vectors.
This decision allows us to 
directly control the text processing 
and ensures consistency with our approach
to the Math Genealogy Project dataset,
where textual data 
is employed to model the
attributes
probabilities.

\subsection{Learning task}

Each paper within the graph 
is categorized into one of 40 designated 
subject areas,
as determined by both the
authors and the moderators at arXiv.
The goal of this learning task is
to predict the subject area of each paper,
making it a 40-class classification problem. 
The nodes are split into three subsets: 
training, validation, and testing. 
The training nodes are used to 
estimate the models parameters, 
the validation nodes are utilized for
hyperparameter tuning,
and the testing nodes are reserved for 
evaluating the final performance
of the trained models. 
For this benchmark, 
the dataset is split based 
on the publication year of each paper.
To evaluate model performance,
the accuracy metric is used, 
assessing how well the models
classifies papers into their correct subject areas.

\subsection{Application of the model}

In this section, we detail the process of
adapting the model to the ogbn-arxiv dataset.
This involves estimating 
the relevant parameters and 
outlining the statistical assumptions 
made to compute the predictions.

\subsubsection{Estimating the functions $\psi_i$ and $\phi_i$}

To model the conditional probabilities $\psi_i$ and $\phi_i$,
we propose a 
parametric distribution.
The estimation of the respective parameters
is performed using the samples of in-degrees and out-degrees,
considering only the training nodes.
However, for each training node,
both in-degree and out-degree are computed
by considering all nodes in the graph, not just the training set.

The functions $\psi_i$
model the in-degree of a node, 
representing the number of citations a paper receives.
For these probabilities, 
we propose 
a parametric probability mass function
parameterized by
$\beta_i \in (0, 1)$,
$\mu_i \in \mathbb{R}$,
and $\sigma_i > 0$.
Here, $\beta_i$ defines the probability
that a paper with label $i$
receives no citations.
Additionally, the parametric function 
follow a discrete log-normal distribution
with parameters $\mu_i$ and $\sigma_i$ 
when restricted to the positive integers. 
The probability mass function is given by:

\begin{equation}\label{psi_in_degree_arxiv}
    \psi_i (d; \beta_i, \mu_i, \sigma_i)
    =
    \begin{cases}
    \beta_i 
    & \text{ if } d = 0, \\
    (1 - \beta_i)
    \frac{(\sigma_i d)^{-1}
    \text{exp}\left( - (\text{ln}(d) - \mu_i)^2 
    (2 \sigma_i^2)^{-1} \right)}
    {\sum_{d'=1}^{\infty}
    (\sigma_i d')^{-1}
    \text{exp}\left( - (\text{ln}(d') - \mu_i)^2 
    (2 \sigma_i^2)^{-1} \right)}
    & \text{ if } d \geq 1. \\
    \end{cases}
\end{equation}

\begin{figure}[ht]
    \centering
    \includegraphics[width=0.99\linewidth]{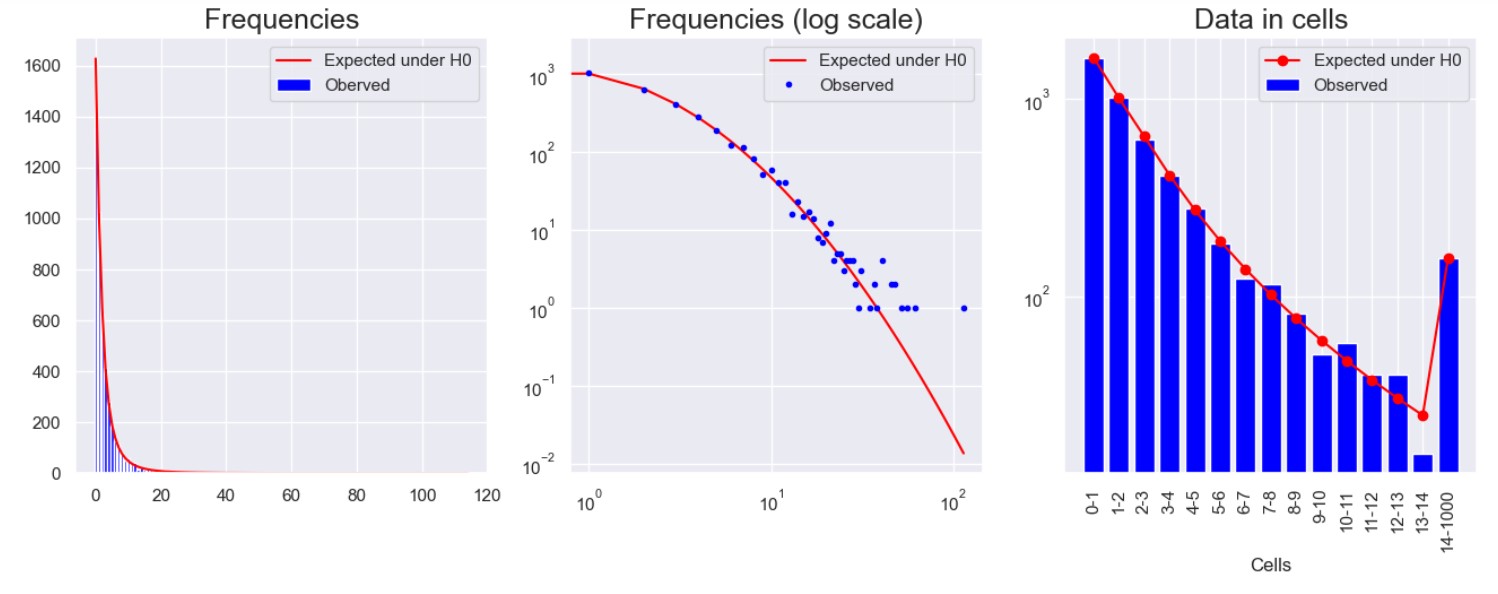}
    \caption{Chi-squared goodness-of-fit for the in degree of nodes with label 2.
    For this sample, the test statistic
    takes the value $T = 14.36$, 
    yielding a p-value of 0.21.
    Therefore, the test concludes that
    the data
    follows the proposed distribution.}
    \label{fig:arxiv_in_2}
\end{figure}

The estimation $\hat{\beta_i}$
is calculated as the proportion 
of papers with label $i$ that have no citations.
The parameters $\hat{\mu_i}$ and $\hat{\sigma_i}$
are then estimated via maximum likelihood
using a computational program.
To validate the suitability
of the proposed parametric functions,
we conduct goodness-of-fit hypothesis tests,
following the same procedure as in Section \ref{estimacion_phi_mgp}.
Specifically, for each label $i \in \mathcal{Y}$,
we perform a 
Chi-squared goodness-of-fit test with $k = 15$ cells,
or less cells if the expected values are not valid,
to evaluate whether the in-degree distribution
fits the proposed model.
An example of 
goodness-of-fit hypothesis test
for the in-degree of label 2
is presented in Figure \ref{fig:arxiv_in_2}.
Out of the 40 labels, 
34 follow the proposed distribution.
For the remaining 6 labels, 
the in-degree distribution does not fit perfectly.
However, previous work in scientometrics
has shown that the probability of an article being cited
can be well approximated by a log-normal distribution 
\cite{lognormal_citation, discrete_lognormal_citation_data}.
Thus, we proceed with using Equation \ref{psi_in_degree_arxiv}
to model the distribution of in-degree
conditioned on the labels.

A similar approach is applied 
to model the conditional probabilities $\phi_i$,
which govern the distribution of the out-degree.
We propose parametric functions
$\phi_i (d; \beta_i', \mu_i', \sigma_i')$,
of the same form of Equation  \ref{psi_in_degree_arxiv}.
As with the in-degree,
the parameter $\beta_i'$ 
defines the probability that
a paper with label $i$ does not cite any other paper,
and 
the proposed out-degree distribution 
follows a discrete log-normal distribution,
parameterized by $\mu_i'$ and $\sigma_i'$ 
when restricted to the positive integers.
To validate these parametric functions, 
we perform a
Chi-squared goodness-of-fit test with $k = 15$ cells
for each label $i \in \mathcal{Y}$, as described earlier.

However, the results for the out-degree distribution 
are less favorable compared to the in-degree. 
Specifically, only 23 out of the 40 labels
pass the goodness-of-fit test.
Given these results,
the proposed parametric functions may not be well-suited for modeling out-degree distributions in this dataset. 
As a more flexible alternative,
we also explore non-parametric estimation 
based on empirical frequencies, 
as described in Equation \ref{estimar_phi}, 
where we set $D_{\phi} = 1000$.
We maintain both the parametric functions 
and the frequency-based method,
treating the choice of method as a hyperparameter,
which
is tuned using
the validation set.

\subsubsection{Estimating other parameters}

To estimate the remaining parameters,
we follow the same procedure used 
for the Math Genealogy Project dataset.
Specifically, the transition probabilities 
$\Theta$, $\Xi$, 
and the prior probabilities $\pi$
are estimated using additive smoothing,
with hyperparameters 
$\alpha_{\pi},$
$\alpha_{\Theta},$
and $\alpha_{\Xi}$ 
optimized via the validation set.
Since the raw text (title and abstract)
serves as the node attributes,
we estimate the conditional probabilities
$\omega_i$ 
according to the methodology described
in Section \ref{parameter_estimation}.
The vocabulary $\Sigma$ is constructed 
by extracting all unigrams and bigrams 
from the training nodes' text, excluding stop words.
In line with Section \ref{estimacion_omega_mgp},
we also experiment with an 
alternative vectorization method using TF-IDF.
For the experiments,
both vectorization approaches 
(count and TF-IDF)
are considered, 
and the choice is treated as a hyperparameter,
selected based on performance in the validation set.

\subsection{Results on the ogbn-arxiv dataset}

This section presents the selected hyperparameters 
for our probabilistic model,
and the resulting classification performance
on the test set.
We compare our model against 
four benchmark methods reported
in the original paper proposing 
this dataset \cite{ogbn_arxiv_paper}.

For our model, 
hyperparameters were tuned on the validation set
maximizing the accuracy score.
Following the same approach 
used for the Math Genealogy Project dataset,
we evaluate the MAP and ML estimation 
using shared parameters and hyperparameters.
We consider four iterations 
to compute predictions,
selecting the one with best performance on validation data.
Table \ref{hyper_ogbn} summarizes the selected hyperparameters.

\begin{table}[ht]
    \begingroup
    \renewcommand{\arraystretch}{0.9} 
    \centering
    \begin{tabular}{@{} l c @{}}
    \toprule
        \textbf{Hyperparameter} & \textbf{Selection} \\ 
    \midrule
        Text vectorization method & TF-IDF \\
        N-gram range & Unigrams and Bigrams (1-2)  \\  
        Minimum document frequency & 1 \\ 
        Maximum document frequency & 0.5 \\ 
        Max features in vocabulary & No limit \\ 
        $\alpha_{\omega}$ & 0.002 \\
        $\alpha_{\pi}$ & 0 \\ 
        $\alpha_{\Theta}$ & 1 \\ 
        $\alpha_{\Xi}$ & 1 \\ 
        Distribution of $\phi_i$ & Equation \ref{psi_in_degree_arxiv} \\  
        Estimation at iteration 0 & Equation 
        \ref{iteracion_0_texto}\\
        Best iteration ML &   2 \\
        Best iteration MAP &   2 \\
    \bottomrule
    \end{tabular}
    \caption{Selected Hyperparameters for the ogbn-arxiv.}
    \label{hyper_ogbn}
    \endgroup
\end{table}

In Table \ref{resultados_ogbn}
we present the accuracy scores 
for our probabilistic model 
on the validation and test sets,
comparing it to benchmark methods from 
\cite{ogbn_arxiv_paper}. 
These benchmarks use t
he 128-dimensional node features 
and treat the graph as undirected,
as opposed with our model that considers the raw
text data as attributes, and a directed graph. 
As anticipated,
the Multi-Layer Perceptron (MLP) model,
which does not incorporate graph structure,
exhibits the lowest accuracy.
On the other hand, 
methods that leverage graph information
show improved results.
Our probabilistic model, 
both with ML and MAP estimations,
outperforms all these baselines,
achieving the highest accuracy on both 
the validation and test sets.
Specifically, the MAP estimation achieves 
the best performance.
These results demonstrate 
the competitive advantage 
of our probabilistic approach
in accurately classifying nodes 
in the ogbn-arxiv dataset.

\begin{table}[ht]
    \begingroup
    \renewcommand{\arraystretch}{0.9} 
    \centering
    \begin{tabular}{@{} l c c @{}}
    \toprule
        \textbf{Method} & \textbf{Validation Accuracy} & \textbf{Test Accuracy} \\ 
    \midrule
        Our Model (ML)      & 0.7573            & 0.7403 \\ 
        Our Model (MAP)     & \textbf{0.7586}   & \textbf{0.7432} \\ 
        MLP                 & 0.5765            & 0.5550 \\ 
        Node2Vec \cite{node2vec} & 0.7129      & 0.7007 \\ 
        GCN \cite{gcn_original}   & 0.7300      & 0.7174 \\ 
        GraphSage \cite{graphsage}& 0.7277     & 0.7149 \\ 
    \bottomrule
    \end{tabular}
    \caption{Results for Testing Data in ogbn-arxiv. The baselines are obtained from \cite{ogbn_arxiv_paper}.}
    \label{resultados_ogbn}
    \endgroup
\end{table}

\section{Conclusion}

In this work, 
we presented a probabilistic model 
for node classification in directed 
attributed graphs. Our findings demonstrate that the proposed model
not only provides interpretable predictions,
but also achieves competitive performance
compared to existing benchmarks.
The interpretability and efficiency 
of our model is a significant advantage
against other node-classification methods.
Additionally, we introduced a new dataset
for node classification and implemented
several learning algorithms in this dataset
that can serve as benchmarks. 
This enhances the resources available for researchers in this domain.

One promising direction for 
future research is the application of our model
to graphs that do not satisfy the homophily property.
Given that our model can determine transition probabilities,
it may effectively learn scenarios where connected nodes
are likely to have different labels.
This characteristic could lead to valuable insights 
and satisfactory predictive performance in such graphs.
Additionally, the model could be further generalized
to consider information beyond the first-order 
neighborhood of a node for the predictions,
which could improve its predictive performance.

\bibliographystyle{plain} 
\bibliography{references}

\end{document}